\documentclass[conference]{IEEEtran}
\IEEEoverridecommandlockouts
\usepackage{cite}
\usepackage{amsmath,amssymb,amsfonts}
\usepackage{graphicx}
\usepackage{textcomp}
\usepackage{xcolor}
\usepackage{algorithm}
\usepackage{algpseudocode}
\usepackage{tabularx}
\usepackage{subfigure}
\usepackage{multirow}
\usepackage{enumerate}  

\usepackage[algo2e]{algorithm2e} 
\usepackage{amsthm}
  {
      \theoremstyle{plain}
      \newtheorem{assumption}{Assumption}
      \newtheorem{theorem}{Theorem}
      \newtheorem{remark}{Remark}
      \newtheorem{proposition}{Proposition}
      
      \newtheorem{definition}{Definition}
  }

\def\BibTeX{{\rm B\kern-.05em{\sc i\kern-.025em b}\kern-.08em
    T\kern-.1667em\lower.7ex\hbox{E}\kern-.125emX}}
\begin{document}

\pagenumbering{arabic}
\title{On the Convergence of Gossip Learning in the Presence of Node Inaccessibility}

\author{\IEEEauthorblockN{Tian Liu\textsuperscript{{\textdagger}1}, Yue Cui\textsuperscript{{\textdagger}2}, Xueyang Hu\textsuperscript{3}, Yecheng Xu\textsuperscript{4}\thanks{\textsuperscript{{\textdagger}} Equal contribution.}, Bo Liu\textsuperscript{5}}
\IEEEauthorblockA{\textsuperscript{1}Zhejiang Laboratory,
Hangzhou, China \\
\textsuperscript{2} Department of Computer Science and Computer Information Systems, Auburn University at Montgomery, Montgomery, USA\\
\textsuperscript{3}Department of Computer Science and Software Engineering, Auburn University, Auburn, USA \\
\textsuperscript{4}Institute of Rural Development, Zhejiang Academy of Agricultural Sciences, Hangzhou, China\\ 
\textsuperscript{5} DBAPPSecurity Co., Ltd., Hangzhou, China\\
 tianliu@zhejianglab.com, ycui1@aum.edu, xueyang.hu@auburn.edu, 
 xyc2022@zaas.ac.cn, bo.liu@dbappsecurity.com.cn}
}

\maketitle

\begin{abstract}
Gossip learning (GL), as a decentralized alternative to federated learning (FL), is more suitable for resource-constrained wireless networks, such as Flying Ad-Hoc Networks (FANETs) that are formed by unmanned aerial vehicles (UAVs). GL can significantly enhance the efficiency and extend the battery life of UAV networks. Despite the advantages, the performance of GL is strongly affected by data distribution, communication speed, and network connectivity. However, how these factors influence the GL convergence is still unclear. Existing work studied the convergence of GL based on a virtual quantity for the sake of convenience, which failed to reflect the real state of the network when some nodes are inaccessible. In this paper, we formulate and investigate the impact of inaccessible nodes to GL under a dynamic network topology. We first decompose the weight divergence by whether the node is accessible or not. Then, we investigate the GL convergence under the dynamic of node accessibility and theoretically provide how the number of inaccessible nodes, data non-i.i.d.-ness, and duration of inaccessibility affect the convergence. Extensive experiments are carried out in practical settings to comprehensively verify the correctness of our theoretical findings. 

\end{abstract}


\section{Introduction}
\label{sec::intro}
Federated learning (FL), as a collaborative learning method, has thrived in many areas, but the need of a central server hinders its applications in wireless resource-constrained scenarios. One typical scenario 
is smart agriculture, in which unmanned aerial vehicles (UAVs) play important roles in crop scouting\cite{qu2022uav}, field coverage mapping\cite{albani2019field}, plant protection \cite{yu2022key}, aquaculture water quality monitoring \cite{davis2023developing}, and forest regeneration \cite{mohan2021uav}. By incorporating collaborative learning into the above application, parallel monitoring and prediction tasks, such as crop health analysis and harvest prediction, can be achieved while performing regular irrigation and fertilization operations. However, energy management remains a significant concern. Given that agricultural UAVs often operate in rural areas, which are far from base stations, employing FL on UAVs could rapidly drain the battery, making FL less feasible for such applications.

Gossip learning (GL), as a decentralized learning algorithm, is an alternative to FL. GL can overlay the Mobile Ad-Hoc Network (MANET) or the Flying Ad-Hoc Network (FANET) to enable decentralized training in an infrastructure-less network layout based on opportunistic physical contacts. The decentralized framework provided by GL can greatly improve the learning efficiency and maximize the battery life of UAVs. 




GL differs from FL in how model aggregation is performed. Specifically, the model aggregation in GL is performed in a self-organized way, in which a pair of nodes exchange model updates if and only if they are connected by a direct communication link. GL is highly suitable for network configuration with low bandwidth and high latency \cite{lian2017can} and the advantages of GL over FL are mainly fourfold. First, more energy-efficient communication protocols, such as Wi-FI, can be used. According to the specification of DJI Agras T16, the long distance transmission protocol (OcuSync 2.0) consumes 4 times the power of 802.11x. Second, the pairwise data sharing scheme does not rely on the full network topology. This fits FANETs where the network topology changes fast, hence nodes may not have the full network topology all the time. Third, GL is robust to single-point failure, latency, and collision by the redundancy introduced by pairwise data sharing. Last, GL achieves a comparable or better performance under compression compared with FL \cite{hegedHus2019gossip, hegedHus2021decentralized}.

Despite the advantages above, the performance of GL is greatly affected by data distribution, communication speed, and network connectivity \cite{giaretta2019gossip}. It has been shown that GL maintains the original convergence speed only in network with good connectivity, independent and identically distributed (i.i.d.) data and homogeneous communication speed, otherwise slow convergence or biased model could happen. However, the study of how each factor affects convergence is still in its infancy. 

\begin{figure}[ht!]
\centering
\includegraphics[width=0.9\linewidth]{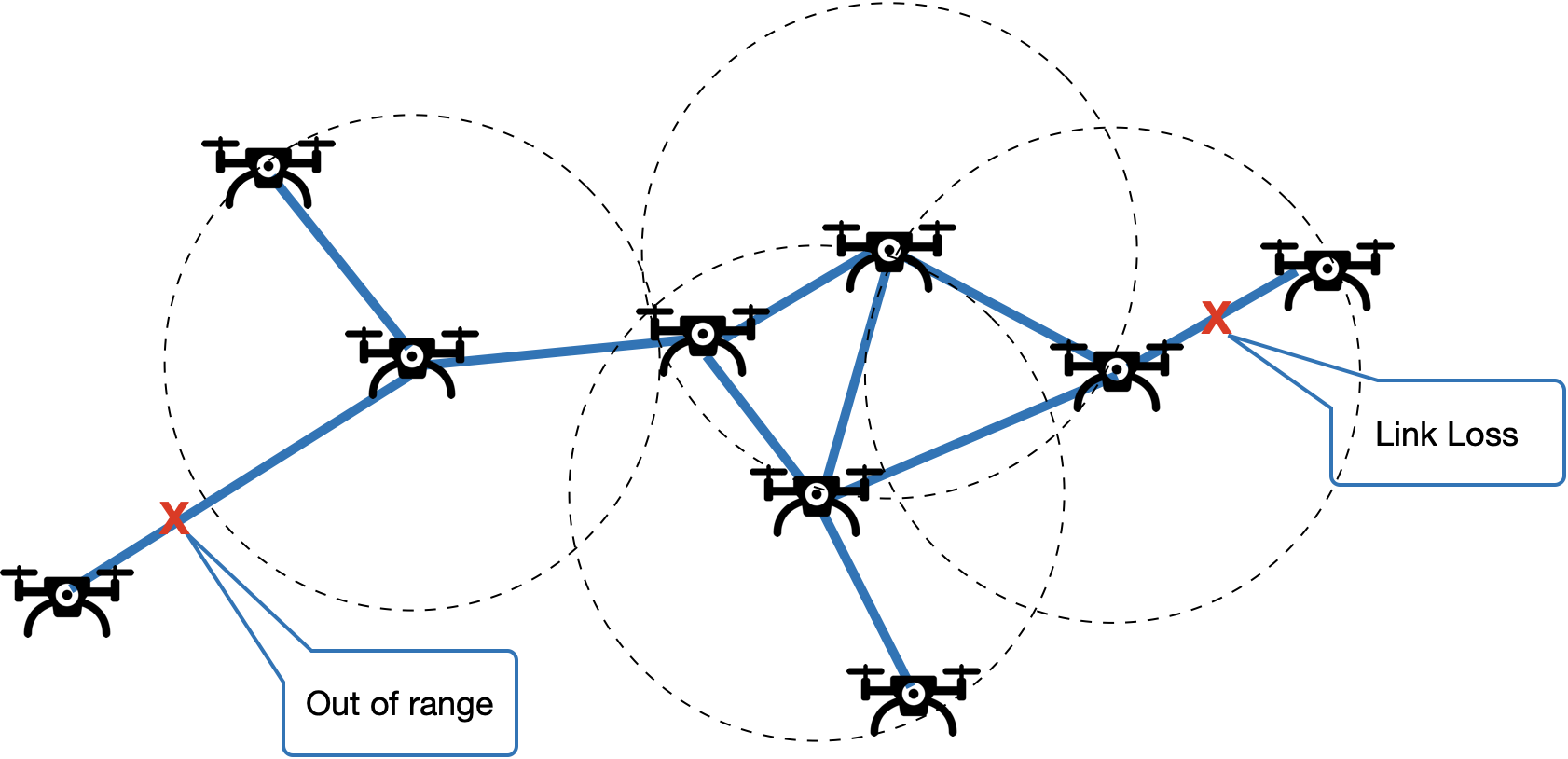}
\caption{The two scenarios that the average of all local models fails to reflect the GL state of the network.}
\label{fig::example}
\end{figure}

Recently, efforts have been made to study the convergence behavior of GL in networks such as MANETs and FANETs \cite{koloskova2019decentralized, tang2022gossipfl, hashemi2021benefits}. Since gossip protocol is an opportunistic protocol, for convenience, all the aforementioned works studied the convergence of GL based on a virtual quantity, which is the average of all local models. However, the use of this virtual quantity relies on the assumption that the gossip exchange preserves the average of the network. As shown in Fig. \ref{fig::example}, nodes may be inaccessible due to out-of-range or link loss, causing failure in gossip exchange. In the presence of such nodes, the virtual average does not reflect the real state of the network. Although GL is known for its ability to handle single-point failures, such failures may negatively affect in subsequent training rounds. On one hand, inaccessible nodes keep learning on its local data may deviate from others and introduce bias. On the other hand, when the inaccessible nodes rejoin, the outdated local model integrated in the current gossip exchange would slow down the convergence. 

Our work aims to investigate the impact of inaccessible nodes under dynamic network topology from the perspective of convergence analysis. To our best knowledge, this work offers the first theoretical investigation into the convergence of GL considering inaccessible nodes. Our contribution in this study are threefold. 
\begin{itemize}
    \item We find that the accessible and inaccessible nodes present different learning patterns. Furthermore, we find that the weight divergence between the average of all local models and the average considering the inaccessibility is related to both data non-i.i.d.-ness and the number of inaccessible nodes.
    \item We investigate the impact of non-i.i.d.-ness, the duration of inaccessibility, the number of inaccessible nodes on the GL convergence under the dynamic of node inaccessibility. More importantly, our findings show that the performance gap to the optimal solution is an increasing function of the number of inaccessible nodes. Implications based on theoretical findings are also provided.
    \item We conducted extensive numerical experiments to validate our findings, considering a comprehensive set of parameters, including the dropout rate, non-i.i.d.-ness across local data, and inaccessible duration from real agricultural UAV applications.
\end{itemize}


Throughout this paper, we use the notation in Table \ref{tab::notation}.

\begin{table}[ht]
\caption{Notation and definitions.}
\begin{center}
\begin{tabular}{>{\centering}p{2cm} p{6cm}}
\hline
\textbf{Notation} & \textbf{Definitions}\\ 
\hline
 $\| \cdot \|$ &  $\ell_{2}$ norm.\\
 $D_i$, $D$ & The data on the $i$-th node and the aggregation of all data, respectively. $D = \cup_{i=1}^n D_i$.\\
 $\mathcal{S}, \mathcal{A}(t)$ & The set of all nodes and the set of accessible nodes at time $t$, respectively. Therefore, the set of inaccessible nodes is denoted by $\mathcal{S/A}(t)$.\\
 $\tau(t,i)$ & The number of inaccessible rounds of device $i$ at time $t$.\\
 $n_1(t)$, $n_2(t)$, $n$ & The number of nodes in $\mathcal{A}(t)$, $\mathcal{S}/\mathcal{A}(t)$, and $\mathcal{S}$, respectively. $n= n_1(t) + n_2(t)$. \\
 $w_i$,  $\bar{w}$, $\widetilde{w}$ & The $i$-th local model, the average of all local models, and the average considering inaccessibility, respectively. \\
  $F_i(\cdot)$, $F(\cdot)$ & Local objective on the $i$-th node and global objective, respectively.\\
\hline
\end{tabular}
\label{tab::notation}
\end{center}
\end{table}

\section{Related Work}
Gossip protocol \cite{boyd2006randomized} is an opportunistic protocol, originally designed to maintain consistency across a decentralized network, especially a dynamic network, and to cope with single-point failure. In particular, every node periodically sends out the new message to a subset of its neighbor nodes. Eventually, the entire network will receive the particular message with a high probability by random walk theory. Scholars in \cite{ormandi2013gossip} extended it to gossip learning and applied it to decentralized linear models due to privacy considerations. Then various learning schemes were explored, such as deep learning \cite{blot2016gossip}, reinforcement learning \cite{mathkar2016distributed}, and recommendation systems \cite{belal2022pepper}. It was claimed in \cite{hegedHus2019gossip, hegedHus2021decentralized} that GL could achieve similar or better performance compared to FL. There is another line working on the scalability and efficiency. To overcome the communication bottleneck, scholars \cite{koloskova2019decentralized} proposed a gossip algorithm with compressed communication and analyzed the convergence for biased and unbiased compression operators. Tang et al. proposed a sparsification algorithm\cite{tang2022gossipfl}, and Hashemi et al. focused on the impact of the number of gossip steps and studied the convergence for a non-convex objective \cite{9664349}. Yuan et al. analyzed convergence of GL under convex objective and fixed learning rate \cite{yuan2016convergence}. Since gossip protocol is an opportunistic protocol, all the aforementioned works studied the behavior of GL in terms of the average of all local models for convenience, which relies on the assumption that the gossip averaging preserves the average. However, in the presence of inaccessible nodes, this quantity does not represent the state of the whole network. Therefore, the impacts of connectivity, data heterogeneity, number of inaccessible nodes, and duration are still unclear. In summary, the understanding of the convergence behavior of GL is still in its infancy. 

\section{Problem Setup and Assumptions}
\subsection{Gossip Learning}
Consider a scenario where $n$ devices collaboratively train a model with parameter $w \in \mathbb R^d$. Each node $i,~i \in [n]$ preserves a local model $w_i$ and has its own data $D_i$. And we wish to minimize an objective function $F(w)$ 
\begin{small}
\begin{flalign}
     F(w): =  \frac{1}{n} \sum_{i=1}^n F_i(w),
\end{flalign}
\end{small}
where $F_i(\cdot)$ is the loss function defined by the data available on each node. 

Conventional FL training consists of local training and server aggregation steps. GL is the same as FL in the local training step, and the main difference lies in the aggregation step. Since the central server is absent in GL, the model sharing in GL is performed by a direct communication link between a pair of nodes, or neighbors. The model aggregation is then performed locally based on the shared models \cite{koloskova2019decentralized}. GL is formally defined in \cite{yuan2016convergence}. 
\begin{enumerate}

    \item \textbf{Gossiping average.} Each node $i$ computes the weighted average by the gossiping matrix $G$, where its element $g_{ij} \neq 0$ only if $j$ is a neighbor of $i$ or $j = i$:
    \begin{small}
    \begin{flalign}
        w_i^{t+\frac{1}{2}} &= \sum_j g_{ij}^t  w_i^t.
    \end{flalign}
    \end{small}
    \item \textbf{Local training.} Each node $i$ updates its own local model $w_{i}$ by running an stochastic gradient descent (SGD) on the local dataset $D_i$ and calculate the gradient:
    \begin{small}
    \begin{flalign}
        \Delta w_i^{t+\frac{1}{2}} = \eta \nabla F_i(w^{t+ \frac{1}{2}}_{i}),
    \end{flalign}
    \end{small}
    where $\eta$ is the learning rate, which can be either fixed or varying with $t$.
    \item \textbf{Model update.} Each node applies the calculated gradients to the averaged model:
    \begin{small}
    \begin{flalign}
        w_i^{(t+1)} &= w_i^{t+\frac{1}{2}} - \Delta w_i^{t+\frac{1}{2}}.
    \end{flalign}
    \end{small}
    
\end{enumerate}

Ideally, the nodes in the network are well-connected, and models can be averaged quickly and extensively. However, as we discussed, this does not depict a real-world scenario, especially in wireless and resource constrained case, in which a node may be out of range or experiencing a link loss. This motivates our evaluation of how node inaccessibility would affect the GL performance. We evaluate some basic questions regarding the node inaccessibility in GL:\textbf{ (1) How much bias does the average of all local models introduce to the convergence analysis when there are inaccessible nodes? (2) How does the inaccessible nodes affect the convergence of GL? (3) What variants may be potential to mitigate this bias?}

\subsection{Assumptions and notation}
This paper studies the convergence of GL under the following assumptions and notation.
\begin{definition} \label{def::doubly} The gossip matrix $G = [g_{ij}] \in [0, 1]^{n\times n}$ associated with a connected graph satisfies: (1) If $i\neq j$ and $i, j$ is not connected, then $g_{ij} =0$; otherwise, $g_{ij} >0$ ; (2) $G=G^T$ (symmetric); and (3) $\sum_{i}G_{ij} = \sum_{j}G_{ij} =1$ (doubly stochastic).
\end{definition}

\begin{assumption} 
\label{assumption_smooth}
The loss functions $F_i(\cdot)$ for $i \in [n]$ are all L-smooth; that is, $\forall v, w \in \mathbb{R}^d$,
\begin{footnotesize}
\begin{small}
\begin{flalign}
  F_i(v) - F_i(w) \leq \langle v-w, \nabla F_i(w)\rangle + \frac{L}{2} \|v-w\|^2.
\end{flalign}
\end{small}
\end{footnotesize}
\end{assumption}

\begin{assumption} 
The loss functions $F_i(\cdot)$ for $i \in [n]$ are all $\mu$-strongly convex; that is, $\forall v, w \in \mathbb{R}^d$, 
\begin{footnotesize}
\begin{small}
\begin{flalign}
F_i(v) - F_i(w) \geq \langle v-w, \nabla F_i(w)\rangle + \frac{\mu}{2} \|v-w\|^2.
\end{flalign}
\end{small}
\end{footnotesize}
\end{assumption}

\begin{assumption} 
 \label{assumption:: gradient_bound}
The expectation of the squared $\ell_2$ norm of the stochastic gradients is bounded; that is, 
\begin{footnotesize}
\begin{small}
\begin{flalign}
    \mathbb{E}_\xi \|\nabla F_i(w, \xi)\|^2  \leq G^2,
\end{flalign}
\end{small}
\end{footnotesize}
where $\mathbb{E}_\xi$ denotes the expectation against the randomness of the stochastic gradient.

\end{assumption}


Denote the optimal value for $F(\cdot)$ by $F^*$, and the optimal value for $F_i(\cdot)$ by $F_i^*$. Define $\Gamma$ as a measurement of non-i.i.d.-ness across clients: $\Gamma \overset{\Delta}{=}  \sum_{i=1}^n \frac{n_i}{n} F^*_i -F^* $, where $\Gamma \geq 0$ indicates how non-i.i.d. across the client's data. Note that given a large enough number of data samples on nodes, we have $\Gamma \rightarrow 0$ for i.i.d. data distributions. We use the notation $w_i^t$ to denote the model on node $i$ at time $t$. 

\section{Convergence Analysis}
Nodes in GL may present different learning patterns due to multiple factors, such as network connectivity, data heterogeneity among nodes, the number of SGD performed on each nodes, the number of gossip steps, etc. In the presence of inaccessible nodes, learning on such nodes does not contribute to the averaged model, therefore simply taking the average of all local models leads to biased error in the convergence analysis. Furthermore, in FL, the outdated models are overwritten by the latest models before rejoining the training. In contrast, GL retains and merges the outdated models once they become accessible again. As a result, dropped nodes not only stall the convergence, but also introduce bias to the learned model. Therefore, it is necessary to study the behavior of accessible and inaccessible nodes separately. 
\subsection{Weight Divergence}
\begin{figure}[ht]
\centering
\includegraphics[width=0.9\linewidth]{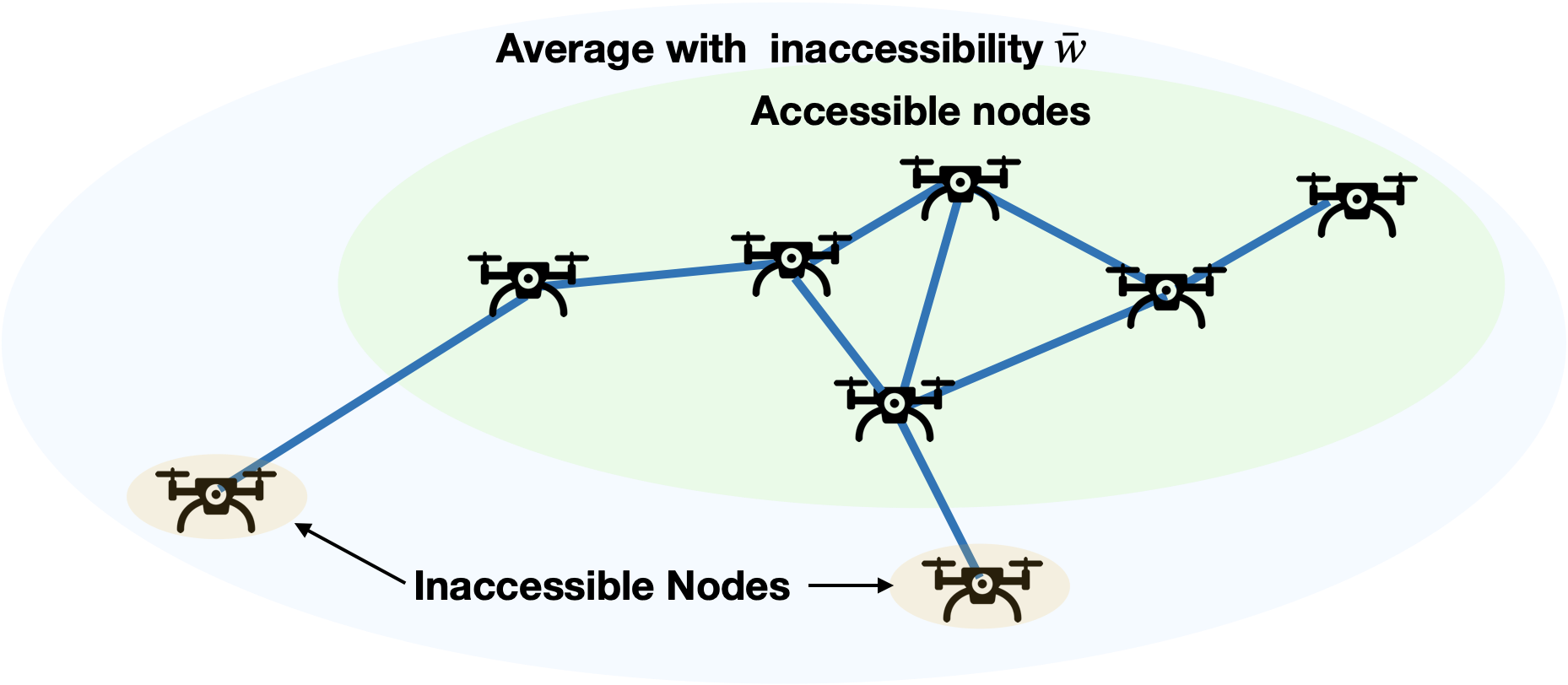}
\caption{Illustration of average with inaccessibility.}
\label{figure::consensus}
\end{figure}

Instead of ignoring inaccessible nodes in other works, we classify the nodes into accessible nodes and inaccessible nodes by whether they successfully transmit the model to their neighbors or not, respectively. To describe both types of nodes, we define the average with inaccessibility, which is illustrated in Fig. \ref{figure::consensus}. 
\begin{definition} \label{def::partial}
 Define the set of accessible nodes at time $t$ as $\mathcal{A}(t)$. For each $i \in \mathcal{A}(t)$, $g_{ii} \neq 1$. And the remaining nodes $i \in \mathcal{S/A}(t)$ are dropped out due to link loss or out of range. Denote the cardinality of $\mathcal{A}(t)$ and $\mathcal{S/A}(t)$ by $n_1(t)$ and $n_2(t)$, respectively. The average with inaccessible nodes is defined as
\begin{small}
\begin{flalign} 
\widetilde{w}^t:= \frac{1}{n_1(t)} \sum_{i \in \mathcal{A}(t)} w_i^t + \frac{1}{n_2(t)} \sum_{i \in \mathcal{S/A}(t)} w_i^t.
\end{flalign}
\end{small}
\end{definition}

The average of all local models are denoted as: $\bar{w}=\frac{1}{n}\sum_{i\in \mathcal{S}} w_i$. The gradient of two averages shows different patterns:
\begin{small} \begin{gather}
g(\bar{w}^{t})= \nabla F_i(\bar{w}^{t}), \\
g(\widetilde{w}^t)=  \frac{n_1(t)}{n}\nabla F_i(\frac{1}{n_1(t)} \sum_{i \in \mathcal{A}(t)} w_i^t) + \frac{1}{n} \sum_{i \in \mathcal{S/A}(t)} \nabla F_i(w_i^t).
\end{gather} \end{small} 

\begin{figure}[ht]
\centering
\includegraphics[width=0.8\linewidth]{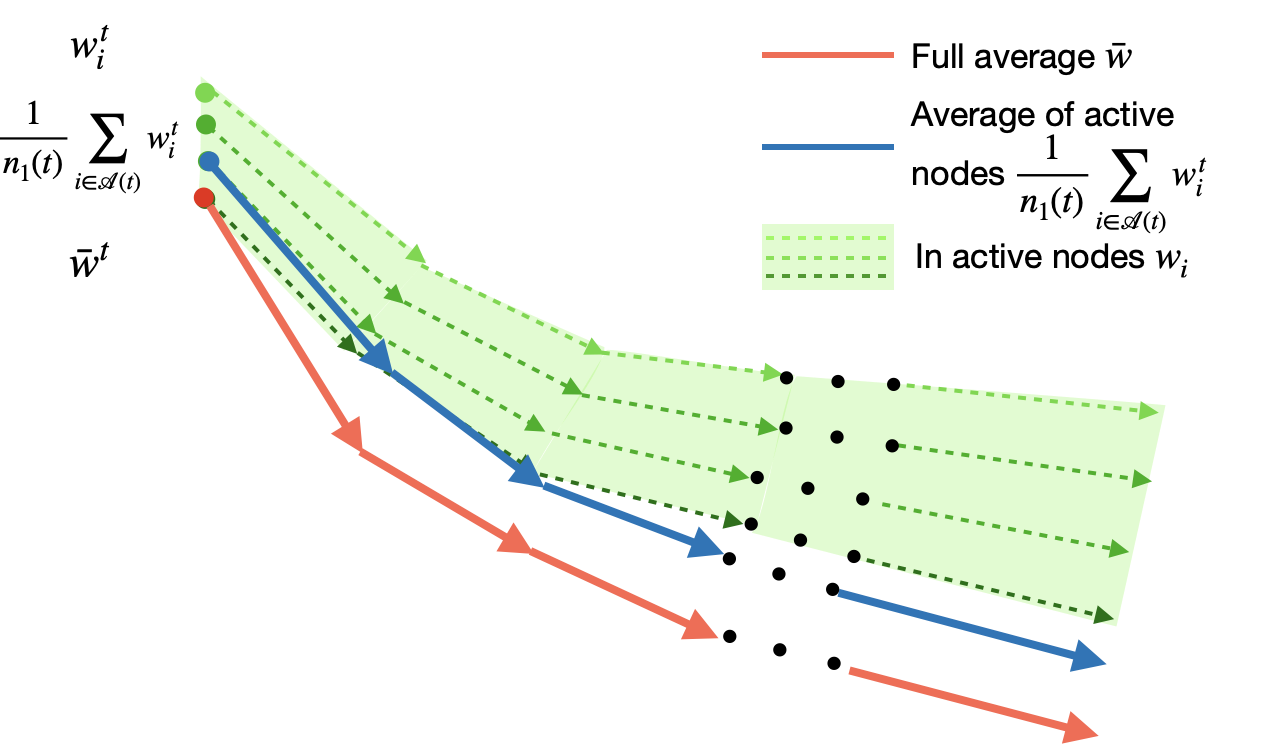}
\caption{Illustration of the weight divergence of full consensus model $\bar{w}$, partial consensus model $\widetilde{w}$ and individual models $w_{i}$.}
\label{figure::weight_divergence}
\end{figure}

$g(\bar{w}^{t})$ in Eq. 6 is simply the gradient on the average of all local models, while $g(\widetilde{w}^{t})$ in Eq. 7 is a weight average of the gradient of the average of all accessible nodes, $ \nabla F_i(\frac{1}{n_1(t)} \sum_{i \in \mathcal{A}(t)} w_i^t)$, and the average gradient of inaccessible models, $\frac{1}{n} \sum_{i \in \mathcal{S/A}(t)} \nabla F_i(w_i^t)$. The descent of the average of all local models, the average of accessible nodes, and inaccessible nodes presents different patterns and can be visualized in Fig. \ref{figure::weight_divergence}. When all nodes are connected (red line), the average is preserved and the local models become more similar after gossip average, thus enjoys the best convergence. However, in the presence of inaccessible nodes, the average on accessible nodes may not converge as fast as in fully connected networks (blue line). In addition, inaccessible nodes training the local model on local data would also introduce divergence (green lines). We give the formal expression of the weight divergence in the following proposition.  

\begin{proposition}
\label{proposition::divergence}
If $F_i$ is $L$-smooth, then we have the following inequality for the weight divergence of between the gradient of the average of all local models $\bar{w}$ and the gradient of average with inaccessibility $\widetilde{w}$:
\begin{small}
\begin{flalign}
&\|g(\widetilde{w}^{t+1})-g(\bar{w}^{t+1})\| \nonumber\\ 
\leq & \frac{L \eta(t)^2 }{n} \Big[\underbrace{n_1(t)(\|\frac{1}{n_1(t)} \sum_{i \in \mathcal{A}(t)} w_i^t - \bar{w}^t \|}_{\text{accessible nodes}}+ \underbrace{\sum_{i \in \mathcal{S/A}(t)} \| w_i^t - \bar{w}^t \|}_{\text{inaccessible nodes}}) \Big].
\end{flalign}
\end{small}

\end{proposition}
The proof can be found in Appendix \ref{proof_divergence}.
\begin{remark}
The first term comes from the accessible nodes, and can be seen as the difference between the sample mean and population mean. Increasing the number of accessible nodes will decrease the first term. The second term comes from the inaccessible nodes. Higher non-i.i.d.-ness across nodes will increase both terms.
\end{remark}

\subsection{Convergence under Node Inaccessibility}
To further analyze the iterative effect of node inaccessibility, we adopt the definition of \cite{gu2021fast} to quantify the dynamics of inaccessible nodes. 

\begin{definition} The inaccessible duration of node $i$ at time $t$ is defined as $\tau(t, i) = t-\max{t'|t'\leq t, i \in \mathcal{A}(t)}$, which is the difference between the current time $t$ and the last time that node $i$ is accessible.
\end{definition}
For convenience, we assume that the inaccessible duration follows an exponential distribution, i.e. $\tau(t, i) \sim \exp(\lambda)$, and we have $\mathbb{E}[\tau(t, i)]= \frac{1}{\lambda}$.
\begin{theorem}
\label{theorem_1}
If the objective function $F_i$ is strongly convex and $L$-smooth, and has bounded gradient, GL satisfies:
\begin{subequations}
\begin{small}
\begin{flalign}
\label{eq_theorem_3}
    &\mathbb{E} \big[\ \|\widetilde{w}^{t+1} - w^*\|^2\big]
    \leq  \alpha(t) ^t \mathbb{E} \big[ \big\|\widetilde{w}^{0} - w^*\big\|^2\big]  + \sum_{i=0}^{t-1}\alpha(t)^i\beta(t)\\
    \alpha(t) =& 2(1- \mu \eta) \\
    \beta(t) = &\frac{1}{n}[\eta L n_1(t) \| \frac{1}{n_2(t)}\sum_{i \in \mathcal{S/A}(t)}w_i^t\|^2 + 4\eta \Gamma + 2n_1(t) \eta^2 G^2 \nonumber \\
    &+ 2n_2(t)G^2(2\eta^3(1+\frac{1}{\lambda}) + \frac{2\mu(1-\eta)}{\lambda})\big]
\end{flalign}
\end{small}
\end{subequations}
\end{theorem}
The proof can be found in Appendix \ref{proof_convergence}.
\begin{remark}
\label{remark_1}
We remark that $\beta(t)$ is an increasing function of the difference between the average of local models of accessible nodes and the average of all local models, i.e. $ \| \frac{1}{n_2(t)}\sum_{i \in \mathcal{S/A}(t)}w_i^t\|^2$, indicating the difference in the last iteration is inherited. Furthermore, $\beta(t)$ is a decreasing function of $\lambda$. Therefore, longer inaccessibility will slow down the convergence. Finally, we note that the data heterogeneity still plays an important role in the convergence of GL.


\end{remark}

\begin{remark}
\label{remark_2}
The term, $\frac{4\mu n_2(t)G^2}{\lambda}$, in $\beta(t)$ is not scaled by the learning rate $\eta$. Therefore, even for a decaying learning rate, i.e. $\lim_{t \rightarrow \infty} \eta(t) = 0$, we have $\lim_{t \rightarrow \infty} \beta(t) \neq 0$. We highlight that in the presence of node inaccessibility, GL does not converge to its optimal solution, and the gap to the optimal solution is an increasing function of the number of inactive nodes $n_2(t)$.
\end{remark}

\subsection{Practical Implications}
Our findings deliver three significant practical implications:
\subsubsection{Increase the Number of Accessible Nodes.}
$\frac{1}{n_1(t)}\sum_{i \in \mathcal{A}(t)}w_i^t$ can be treated as the sample mean drawn from $w_i^t$, while $\bar{w}$ can be seen as the population mean. The difference $\|\frac{1}{n_1(t)}\sum_{i \in \mathcal{A}(t)}w_i^t -\bar{w}^t \|$ is related to the variance of the local models, which originates from the heterogeneity of the data and the sample size, $n_1(t)$. Improving connectivity can reduce this difference. Our findings also help explain why the critical consensus distance'' is a factor of convergence in \cite{zhu2022topology}. 

\subsubsection{Reduce Data Heterogeneity}
According to Eq. 10, data heterogeneity affects GL by the gradient of gossiping averaged model descent on the local data, i.e. $\nabla F_i(\bar{w}, D_i)$, and inaccessible nodes train on their local data individually, i.e. $\nabla F_i(w_i, D_i)$. More non-i.i.d. local distributions among nodes will result in higher divergence in both terms. Applying a drift correction can help preserve distributional stability after averaging. Also, following the intuition that the data of some nodes are more critical than others, it is beneficial to identify those nodes and involve them more frequently.

\subsubsection{De-emphasize the model update of inaccessible nodes when they rejoin} We also notice that the longer the node stays inaccessible, the greater the weight divergence. To mitigate the weight divergence of inaccessible nodes that are back online, one could de-emphasize the outdated model in the averaging operation.


\section{Numerical Experiments}
\subsection{Experimental Setup}
\subsubsection{Datasets and Models}
Model architecture ResNet-18 \cite{he2016identity} is employed in our experiments. Models are trained on CIFAR-10 dataset \cite{krizhevsky2009learning}. The dataset contains 60,000 $32\times 32$ color images in 10 classes, with each class containing 5000 training images and 1000 test images. 

\subsubsection{Training setting}
We report the training parameters in the Tab. \ref{tab::hyper_parameter}. We also introduce two data heterogeneity settings. The first one is \textit{i.i.d. assignment}, in which each node is assigned $600$ images, and the data distribution on each node is the same as the distribution of the entire dataset. The second one is \textit{non-i.i.d. assignment}, in which the nodes do not have the same amount of data and the data distribution on each node is different from the whole dataset. We use Dirichlet distribution with a hyper-parameter $\alpha \in \{1,~10\}$ to generate different data distributions for users, where a smaller $\alpha$ indicates a greater non-i.i.d.-ness. The i.i.d. setting provides the performance upper bound, while the non-i.i.d. setting represents practical scenarios, providing the insight of how non-i.i.d.-ness downgrades the performance in real applications. 
\begin{table}[htbp]
\centering
\caption{Hyper-Training Parameters}
\begin{tabular}{ c c c c c }
\hline
 Epoch & Nodes &Batch size & Learning rate& Local epochs \\ 
\hline
 50 & 14 &128 & 0.1&2 \\  
\hline
\end{tabular}
\label{tab::hyper_parameter}
\end{table}

\subsubsection{Node Movement Simulation}
To simulate the device movement in a dynamic network, we use the Random Waypoint model \cite{MAO2010201}. Each node holds their own dataset $D_i$ and local model $w_i$. The operation area is $1000\:m\times1000\:m$. The trajectories of nodes are generated by the following steps:
\begin{enumerate}[i)]
    \item The node $i$ assumes an initial location $(x_i^0, y_i^0)$. 
    \item Each node moves randomly with a speed $v_i$ chosen uniformly from an interval $[5\:m/s,~7\:m/s]$. 
    \item The node pauses $1\:s$ for $t_i$ after reaching $(x_i, y_i)$.
    \item $i = i + 1$ and go to step ii). 
\end{enumerate}

The communication radius to $r=250\:m$. We further set the probability that a node fails communicate to its neighbors to $p \in \{0\%,~10\%,~20\%\}$. When a node becomes inaccessible, the duration of inaccessibility is drawn from an exponential distribution with parameter $\lambda \in \{0.2,~0.3,~0.5\}$. That is, the node being inaccessible at time $t$ will rejoin training at time $t+\exp(\lambda)$. And larger $\lambda$ indicates shorter inaccessible duration. The connectivity of out-of-range and inaccessible nodes is set to 0, and 1 otherwise. Finally, the connectivity matrix is transformed into a doubly matrix.

\subsubsection{Implementations}
All our experiments are conducted on a computing cluster with NVIDIA A100 Tensor Core GPU @40 GB and Intel Xeon Gold 6138 CPU @2.00 GHz. Node movements are generated by ns-3. The learning process is implemented based on PyTorch. 

\subsection{Results}
\begin{figure}[ht]
\centering
\includegraphics[width=\linewidth]{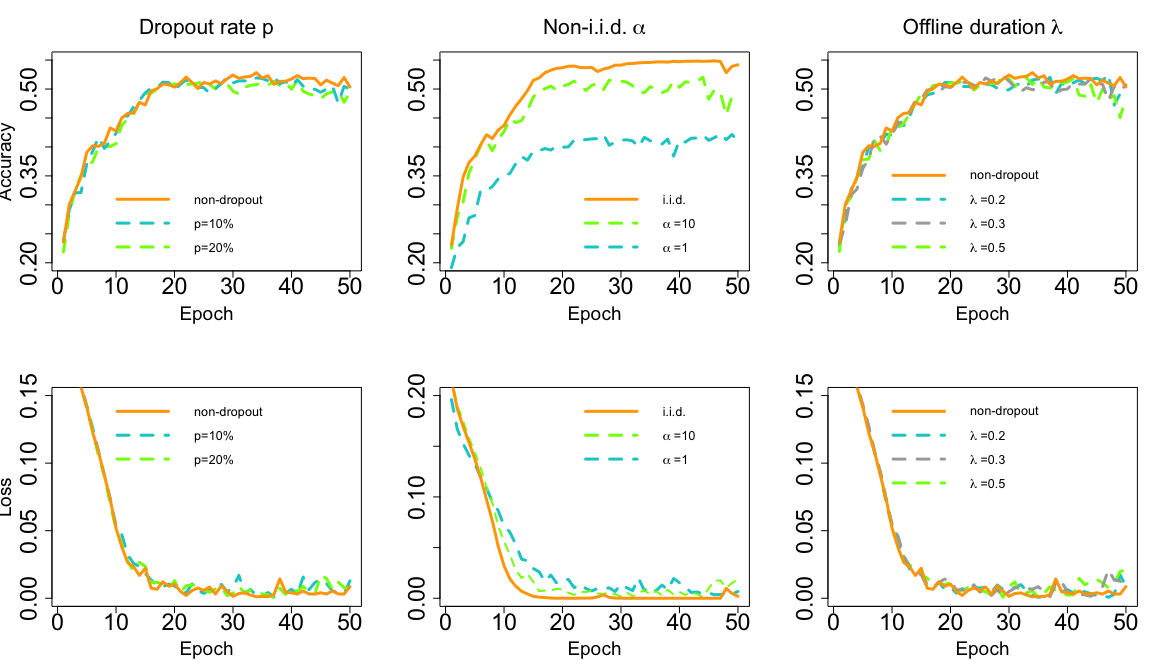}
\caption{The mean test accuracy (first row) and sum of training loss (second row) under different $p,~\alpha,~\lambda$.}
\label{figure::acc_loss}
\end{figure}

We plot the average local test accuracy and sum of training loss as a function of training epoch in Fig. \ref{figure::acc_loss}, considering dropout rate $p$, data non-i.i.d.-ness $\alpha$, and offline duration parameter $\lambda$. The final accuracy and number of inaccessible nodes for each setting are summarized in Table \ref{tab::results}. 
In the first column of Fig. \ref{figure::acc_loss}, we consider the dropout rate $p \in \{0\%,~10\%,~20\%\}$. We fixed other two factors to be $\alpha=10$ and $\lambda=1$. At the beginning, the three curves exhibit identical convergence rates, which is in line with the convergence rate term $\alpha(t)$ in Theorem \ref{theorem_1}. As for the final model accuracy, the non-dropout setting yields the best result, while dropout settings suffer slight performance drops, which differs from the established knowledge that GL is robust against the node dropouts. Moreover, a slight drop in accuracy is observed in dropout settings after convergence is reached. One possible explanation is that frequent changes in participating nodes have an impact on the stability of the convergence.

In the second column of Fig. \ref{figure::acc_loss}, we present the convergence performance under data non-i.i.d.-ness, represented by $\alpha \in {1, 10, \infty}$, with $\infty$ indicating i.i.d.-ness. The i.i.d. setting yields the best convergence, while greater non-i.i.d.-ness leads to reduced accuracy. We can also see that the impact of data non-i.i.d.-ness outweighs the other factors, highlighting that the data heterogeneity remains a big challenge in GL. 

In the third column, we investigate $\lambda \in \{0.2,~0.3,~0.5\}$. In the dropout rate settings, we examine the the impact of multiple node dropouts in one training epoch. In contrast, the $\lambda$ reflects the lagging effect of one node dropout for multiple training epochs. Smaller $\lambda$ indicating a longer inaccessible duration results in a larger performance drop. It can also be seen that the accuracy of the non-dropout case is stable while other three curves experience more fluctuations. The final accuracy of each setting is given in Table \ref{tab::results}.

\begin{table}[htbp]
\centering
\caption{Average Model Accuracy}
\begin{tabular}{ p{0.35 in} p{0.35 in} p{0.45 in} p{0.55 in} c }
\hline
Non-i.i.d $\alpha$ & Dropout rate $p$ & Duration $\lambda$ & Inaccessible nodes & Accuracy ($\%$)  \\ 
\hline
\multirow{4}*{$10$} &  $0\%$ & \multirow{4}*{$1$} &  $1.30\pm 0.81$ & 52.82$\pm$5.45  \\  
 &  $10\%$ & & $3.52 \pm1.76$ & 49.88$\pm$5.66   \\  
 &  $20\%$            &            &  $5.54\pm 1.87$  & 48.31$\pm$6.18   \\
\hline
$\infty$ &  \multirow{3}*{$5\%$} & \multirow{3}*{$1$} & \multirow{3}*{$2.37\pm 1.48$}& 54.18$\pm$3.87   \\  
$ 10$ &   &  & & 49.38$\pm$7.72   \\  
$1$ &             &             &     & 42.28$\pm$6.62    \\

\hline
\multirow{3}*{$10$} &  \multirow{3}*{$10\%$} &   $0.2$        &     $7.26 \pm 2.58$     & 50.71$\pm$8.04  \\
 &                        &   $0.33$      &  $5.56 \pm 2.27$       &    50.53$\pm$6.58\\
  &                        &   $0.5$      &  $4.60 \pm 2.07$  & 48.85$\pm$7.04      \\
\hline
\end{tabular}
\label{tab::results}
\end{table}
\section{Conclusion}
In this paper, we have investigated the impact of inaccessible nodes under dynamic network topology. We have proved that the number of inaccessible nodes, data non-i.i.d.-ness, and duration of inaccessibility negatively affect the GL convergence. Furthermore, we have also proved that the GL does not converge to the optimum in the presence of inaccessible nodes. The convergence gap in the presence of node inaccessibility can be reduced by improving network connectivity, applying bias correction and node selection algorithms, and de-emphasize the model of inaccessible nodes when they rejoin training. Extensive experiments conducted under practical conditions verified the correctness of our findings.
\bibliographystyle{IEEEtran}
\bibliography{reference}

\newpage
\onecolumn
\appendix

\subsection{Proof of Proposition 1}
\begin{proof}
\label{proof_divergence}
At the $t$-th training rounds, the gradient of $\bar{w}$ and $\widetilde{w}$ are:
\begin{small}
\begin{gather*}
g(\bar{w}^{t})= \nabla F_i(\bar{w}^{t})\\
g(\widetilde{w}^t)=  \frac{n_1(t)}{n}\nabla F_i(\frac{1}{n_1(t)} \sum_{i \in \mathcal{A}(t)} w_i^t) + \frac{1}{n} \sum_{i\in \mathcal{S/A}(t)} \nabla F_i(w_i^t),
\end{gather*}
\end{small}
By assumption 2, we have 
\begin{small}
\begin{flalign}
&\|g(\widetilde{w}^{t+1})-g(\bar{w}^{t+1})\|  \nonumber \\
\leq & \frac{n_{1}(t)}{n} \Big[\|\frac{1}{n_1(t)} \sum_{i\in \mathcal{A}(t)} w_i^t- \bar{w}^t\|+ \eta \| \nabla F_i(\frac{1}{n_1(t)} \sum_{i \in \mathcal{A}(t)} w_i^t) - \nabla F_i(\bar{w}^t)\|\Big] + \frac{n_2(t)}{n} \Big[\sum_{i \in \mathcal{S/A}(t)} \| w_i^t - \bar{w}^t\|+ \eta
 \|\sum_{i \in \mathcal{S/A}(t)} (\nabla F_i(w_i^t) - \nabla F_i(\bar{w}^t))\| \Big]\nonumber \\
\overset{(a)}{\leq}  & \frac{1 + L \eta^2 }{n} \Bigg[n_1(t)(\|\sum_{i \in \mathcal{A}(t)} w_i^t - w^t \|+ \sum_{i \in \mathcal{S/A}(t)} \| w_i^t - w^t\|) \Bigg],
\end{flalign}
\end{small}
\end{proof} 
where $(a)$ is due to Assumption 1.
\subsection{Proof of Theorem 1}
\label{proof_convergence}
\begin{proof}
\begin{small}\begin{align}
& \|\widetilde{w}^{t+1} - w^*\|^2\nonumber \\
\leq&\|\frac{n_1(t)}{n}[\widetilde{w} -\nabla F_i(\frac{1}{n_1(t)} \sum_{i \in \mathcal{A}(t)} w_i^t) - w^* ] + \frac{1}{n} \sum_{i \in \mathcal{S/A}(t)} [w_i^t- \nabla F_i(w_i^t) -w^*]\|^2 \nonumber \\
\overset{(b)}{\leq}& 2\| \widetilde{w}^t - w^*\|^2 - \frac{n_1(t)}{n}[\underbrace{4\eta \langle \nabla F_i(\frac{1}{n_1(t)} \sum_{i \in \mathcal{A}(t)} w_i^t), \widetilde{w}^t - w^* \rangle}_{A_1} + 2\eta^2 \|\nabla F_i(\frac{1}{n_1(t)} \sum_{i \in \mathcal{A}(t)} w_i^t)\|^2] \nonumber \\
 &- \underbrace{\frac{1}{n} [4\eta \sum_{i \in \mathcal{S/A}(t)} \langle\nabla F_i(w_i^t), \widetilde{w}^t - w^* \rangle}_{A_2}   +2\eta^2 \sum_{i \in \mathcal{S/A}(t)}\|\nabla F_i(w_i)\|^2],
\end{align}\end{small}
where $(b)$ is due to $\|a+b\|\leq 2\|a\|^2 + 2\|b\|^2$.
\begin{small}\begin{flalign}
    -A_1 = & - 4\eta \big[\langle  \widetilde{w}^{t}-  \frac{1}{n_1(t)} \sum_{i \in \mathcal{A}(t)} w_i^t + \frac{1}{n_1(t)} \sum_{i \in \mathcal{A}(t)} w_i^t- w^*, \nabla F_i(\frac{1}{n_1(t)} \sum_{i \in \mathcal{A}(t)} w_i^t) \rangle\big]\nonumber \\
    = & -4 \eta \big[\langle \widetilde{w}^{t}-  \frac{1}{n_1(t)} \sum_{i \in \mathcal{A}(t)} w_i^t, \nabla F_i(\frac{1}{n_1(t)} \sum_{i \in \mathcal{A}(t)} w_i^t) \rangle + \langle  \frac{1}{n_1(t)}\sum_{i \in \mathcal{A}(t)}w_i^t- w^*, \nabla F_i(\frac{1}{n_1(t)} \sum_{i \in \mathcal{A}(t)} w_i^t)\rangle\big]\nonumber\\ 
    \overset{(c)}\leq&  -4 \eta\big[F_i(\widetilde{w}^t) -F_i(\frac{1}{n_1(t)} \sum_{i \in \mathcal{A}(t)}w_i^t) - \frac{L}{2} \|\widetilde{w}^t - \frac{1}{n_1(t)}\sum_{i \in \mathcal{A}(t)}w_i^t\|^2+ F_i(\frac{1}{n_1(t)}\sum_{i \in \mathcal{A}(t)}w_i^t) -  F_i(w^*) +\frac{\mu}{2} \|  \frac{1}{n_1(t)}\sum_{i \in \mathcal{A}(t)}w_i^t- w^*\|^2\big] \nonumber \\
    \leq& -4\eta [ F_i(\widetilde{w}^t) - F_i(w^*)] + 2L\eta \|\widetilde{w}^t - \frac{1}{n_1(t)}\sum_{i \in \mathcal{A}(t)}w_i^t\|^2 - 2\mu\eta \sum_{i \in \mathcal{A}(t)}\|w_i^t - w^*\|^2\nonumber \\
    \leq & -4\eta L \|\widetilde{w}^t - w^*\|^2 +  2L\eta \|\widetilde{w}^t - \frac{1}{n_1(t)}\sum_{i \in \mathcal{A}(t)}w_i^t\|^2 - 2\mu\eta \sum_{i \in \mathcal{A}(t)}\|w_i^t - w^*\|^2
\end{flalign}\end{small}
where $(c)$ is due to Assumptions 1 and 2. 


\begin{small}\begin{flalign}
    -A_2 = & \frac{4\eta}{n} \sum_{i \in \mathcal{S/A}(t)} \langle  w^*- \widetilde{w}^t  + w_i^{t-\tau(t, i)} -w_i^{t-\tau(t, i)}, \nabla F_i(w_i^{t-\tau(t, i)}) \rangle\big]\nonumber \\
    %
    =& \underbrace{\frac{4\eta}{n} \sum_{i \in \mathcal{S/A}(t)}\langle {w}_i^{t-\tau(t, i)} -w^t , \nabla F_i({w}_i^{t-\tau(t, i)})\rangle}_{B_1} -  \underbrace{\frac{4\eta}{n} \sum_{i \in \mathcal{S/A}(t)}\langle  {w}_i^{t-\tau(t, i)}-w^* , \nabla F_i({w}_i^{t-\tau(t, i)})\rangle}_{B_2}
\end{flalign}\end{small}

\begin{small}\begin{flalign}
    B_1& \leq  \frac{4\eta}{n} \sum_{i \in \mathcal{S/A}(t)}\| {w}_i^{t-\tau(t, i)} - w^t \|^2 + \frac{4\eta^3}{n} \sum_{i \in \mathcal{S/A}(t)} \|\nabla F_i({w}_i^{t-\tau(t, i)})\|^2\nonumber\\
   & \leq  \frac{4\eta^3n_2(t)}{n\lambda}  \big\|\nabla F_i({w}_i^{t-\tau(j,i)})\big\|^2 +\frac{4\eta^3n_2(t)} {n}  \|\nabla F_i({w}_i^{t-\tau(t, i)})\|^2 \nonumber \\
   & \overset{(d)}{\leq} \frac{4\eta^3n_2(t)} {n} (1+\frac{1}{\lambda}) G^2  
\end{flalign}\end{small}
In the last inequality $(d)$, we assume the inaccessible duration follows an exponential distribution, i.e.,  $\tau(t, i) \sim \exp(\lambda)$, and we have $\mathbb{E}[\tau(t, i)]= \frac{1}{\lambda}$.

\begin{small}\begin{flalign}
    -B_2&  \leq \frac{4\eta}{n}  \sum_{i \in \mathcal{S/A}(t)} \big[\langle  w^* - {w}_i^{t-\tau(t, i)} , \nabla F_i({w}_i^{t-\tau(t, i)})\rangle\big] \nonumber \\
    \overset{(e)} \leq &\frac{4\eta}{n}  \sum_{i \in \mathcal{S/A}(t)} \big[ F_i(w^*)- F_i({w}_i^{t-\tau(t, i)}) - \frac{\mu}{2} \|{w}_i^{t-\tau(t, i)} - w^*\|^2 \big] \nonumber\\
     \leq &\frac{4\eta}{n}  \sum_{i \in \mathcal{S/A}(t)}\big[ F_i(w^*) - F_i^* + F_i^* - F_i({w}_i^{t-\tau(t, i)}) - \frac{\mu}{2} \|{w}_i^{t-\tau(t, i)} - w^*\|^2 \big] \nonumber\\
     = & \frac{4\eta}{n}  \sum_{i \in \mathcal{S/A}(t)} (F_i(w^*) - F_i^*) +\frac{4\eta}{n}  \sum_{i \in \mathcal{S/A}(t)} (F_i^* - F_i({w}_i^{t-\tau(t, i)} )- \frac{2\mu\eta}{n} \sum_{i \in \mathcal{S/A}(t)} \big[\|{w}_i^{t- \tau(t, i)} - w^*\|^2\big] \nonumber\\
     \overset{(f)} \leq & \frac{4\eta n_2(t) }{n}  \Gamma - \frac{2\mu\eta}{n} \sum_{i \in \mathcal{S/A}(t)}  \underbrace{\big[\|{w}_i^{t- \tau(t, i)} - w^*\|^2}_{C}\big],
\end{flalign}\end{small}
where $(e)$ is due to Assumptions 2, and $(f)$ is by the definition of $\Gamma$ and the fact that $(F_i^* - F_i({w}_i)) \leq 0$
\begin{small}\begin{flalign}
    -C &\leq -\|{w}_i^{t- \tau(t, i)} - w_i^{t}\|^2 -\|w_i^{t} - w^*\|^2 - 2 \langle {w}_i^{t-\tau(t, i)} - w^{t}, w^{t} - w^*\rangle\nonumber\\
    &\leq -\|{w}_i^{t-\tau(t, i)} - {w}_i^{t}\|^2 - \|{w}_i^{t} - w^*\|^2 + \frac{1}{\eta} \|{w}_i^{t-\tau(t, i)} - {w}^{t}\|^2 + \eta \|{w}_i^{t} - w^*\|^2 \nonumber\\
    & = -(1-\eta) \|{w}_i^t - w^*\|^2 + (\frac{1}{\eta}-1)\|{w}_i^{t-\tau(t, i)} - {w}_i^{t}\|^2 \nonumber \\
    & \leq -(1-\eta) \big[\frac{G^2}{\eta \lambda}  -  \|{w_i}^t - w^*\|^2\big] 
\end{flalign}\end{small}
This completes the proof.
\end{proof}
\end{document}